\documentclass[english]{article}

\usepackage{geometry}
\usepackage[T1]{fontenc}
\usepackage[latin9]{inputenc}
\usepackage{bm}
\usepackage{amsmath}
\usepackage{amssymb} 
\usepackage[unicode=true,
 bookmarks=false, 
 breaklinks=false,pdfborder={0 0 1},colorlinks=false]
 {hyperref}
\hypersetup{
 colorlinks,citecolor=blue,filecolor=blue,linkcolor=blue,urlcolor=blue}
\usepackage{xcolor,colortbl}
\definecolor{Gray}{gray}{0.85}
\usepackage{enumitem}
\makeatletter

\usepackage{amsthm}
\usepackage{cite}  
\usepackage{comment}
\usepackage[sort,compress]{natbib}
\usepackage{booktabs,mathtools}
\usepackage{graphicx}
\usepackage[linesnumbered,ruled,vlined]{algorithm2e}
\usepackage{algorithmic}
\usepackage{multirow}
\usepackage{dsfont}
\usepackage{color}
\usepackage{float}
\usepackage{wrapfig}
\usepackage{graphicx}
\usepackage{microtype}
\usepackage{subfigure}
\usepackage{booktabs} 
\usepackage{ bbold }
\usepackage{bbm}

\usepackage{multirow}
\usepackage{xspace}
\newcommand{\argmax}{\operatorname*{argmax}}

\newcommand{\dc}{\mathcal{D}}

\newcommand{\qsarsa}{Q_{\mathsf{sarsa}}}
\newcommand{\vsarsa}{V_{\mathsf{sarsa}}}
\usepackage{xcolor,colortbl}
\definecolor{tacell}{RGB}{240,128,128}
\definecolor{blue2}{HTML}{0072BD}
\newcommand{\our}{{\bf Qsarsa-AC}\xspace}
\newcommand{\ourr}{{\bf Qsarsa-AC2}\xspace}

\definecolor{yxc}{RGB}{255,0,0}
\definecolor{yjc}{RGB}{225,0,100}
\definecolor{ytw}{RGB}{255,69,0}
\definecolor{gen}{RGB}{0,0,200}
\definecolor{lxs}{RGB}{138,43,226}
\definecolor{hew}{RGB}{0,47,167}

\definecolor{own_pink}{RGB}{217,25,169}
\definecolor{own_blue}{RGB}{0,100,223}

\definecolor{own_pink}{RGB}{217,25,169}
\definecolor{own_blue}{RGB}{0,100,223}

\allowdisplaybreaks

\newcommand{\defn}{\coloneqq}

\newcommand{\pib}{\pi_{\mathsf{b}}}

\newcommand{\cA}{\mathcal{A}}
\newcommand{\cB}{\mathcal{B}}

\newcommand{\cD}{\mathcal{D}}

\newcommand{\cM}{\mathcal{M}}

\newcommand{\cS}{{\mathcal{S}}}

\newcommand{\mymid}{\,|\,} 

\usepackage{scalerel,stackengine}
\stackMath
\newcommand\reallywidehat[1]{%
\savestack{\tmpbox}{\stretchto{%
  \scaleto{%
    \scalerel*[\widthof{\ensuremath{#1}}]{\kern-.6pt\bigwedge\kern-.6pt}%
    {\rule[-\textheight/2]{1ex}{\textheight}}
  }{\textheight}%
}{0.5ex}}%
\stackon[1pt]{#1}{\tmpbox}%
}
\newcommand\reallywidecheck[1]{%
\savestack{\tmpbox}{\stretchto{%
  \scaleto{
    \scalerel*[\widthof{\ensuremath{#1}}]{\kern-.6pt\bigwedge\kern-.6pt}%
    {\rule[-\textheight/2]{1ex}{\textheight}}
  }{\textheight}%
}{0.5ex}}%
\stackon[1pt]{#1}{\scalebox{-1}{\tmpbox}}%
}

\title{Offline Reinforcement Learning with \\ On-Policy Q-Function Regularization}

 \author{
 	Laixi Shi\thanks{Department of Electrical and Computer Engineering, Carnegie Mellon University, Pittsburgh, PA 15213, USA.}\\
 	CMU 
 	\and
	Robert Dadashi\thanks{Google Research, Brain Team.} \\ 
	Google \\
	\and
	Yuejie Chi\footnotemark[1] \\ 	 
  	CMU
  	\and
	Pablo Samuel Castro\footnotemark[2] \\
 Google  \\
	\and
	Matthieu Geist\footnotemark[2]\\
	Google
 	}

\date{}

\begin{document}

\theoremstyle{plain} \newtheorem{lemma}{\textbf{Lemma}}
\newtheorem{proposition}{\textbf{Proposition}}\newtheorem{theorem}{\textbf{Theorem}} \newtheorem{assumption}{Assumption}

\theoremstyle{remark}\newtheorem{remark}{\textbf{Remark}}

\maketitle

\begin{abstract}

The core challenge of offline reinforcement learning (RL) is dealing with the (potentially catastrophic) extrapolation error induced by the distribution shift between the history dataset and the desired policy. A large portion of prior work tackles this challenge by implicitly/explicitly regularizing the learning policy towards the behavior policy, which 
is hard to estimate reliably in practice.
In this work, we propose to regularize towards the Q-function of the behavior policy instead of the behavior policy itself, under the premise that the Q-function can be estimated more reliably and easily by a SARSA-style estimate and handles the  extrapolation error more straightforwardly. 
We propose two algorithms taking advantage of the estimated Q-function through regularizations, and demonstrate they exhibit strong performance on the D4RL benchmarks.
\footnotetext{This paper is published at European Conference on Machine Learning (ECML), 2023.}

\end{abstract}
\noindent \textbf{Keywords:} offline reinforcement learning, actor-critic, SARSA

\section{Introduction}\label{sec:intro}

\begin{figure}
\centering
\includegraphics[width=0.8\linewidth]{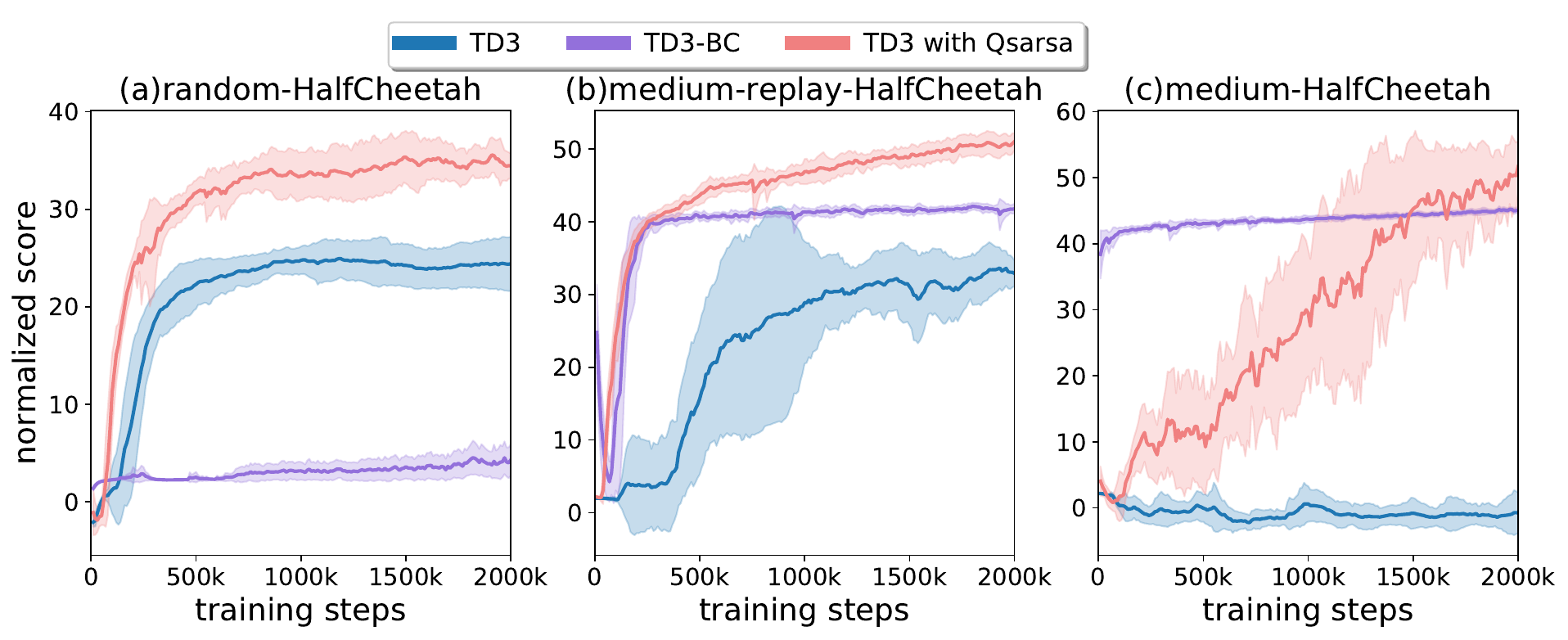}

    \caption{A comparison between our $\qsarsa$-regularized {\sf TD3} method ({\sf TD3 with Qsarsa}), the vanilla {\sf TD3} method from online RL \citep{fujimoto2018addressing},
    and a strong baseline for offline RL {\sf TD3-BC}  \citep{fujimoto2021minimalist} in the tasks of MuJoCo in D4RL.
    }
    \label{fig:intro-motivate}
\end{figure}

Reinforcement learning (RL) has witnessed a surge of practical success recently, with widespread applications in games such as Go game and StarCraft II \citep{silver2017mastering,vinyals2019grandmaster}, control, autonomous driving, etc \citep{arulkumaran2017brief,levine2018reinforcement}. Online RL relies on sequentially collecting data through interactions with the environment.
However, new sample collections or interactions might be infeasible due to privacy, safety, or cost, especially in real-world applications.
To circumvent this, offline/batch RL seeks to learn from an existing dataset, without further interaction with the environment~\citep{levine2020offline}.

The history dataset can be regarded as generated by some unknown behavior policy $\pib$, which is not necessarily of the desired quality or has insufficient coverage over the state-action space. This results in one of the major challenges of offline RL: {\em distribution shift}. Here, the state-action distribution under the behavior policy may heavily differ from that produced by a more desirable policy. As a result, the policy evaluation process presents considerable extrapolation error over the {\em out-of-distribution} (OOD) regions (state-action pairs) that are insufficiently visited or even unseen in the 
history dataset. To address the extrapolation error, prior works mainly follow three principles: 1) {\em behavior regularization:} regularizing the learning policy to be close to the behavior policy or to imitate the (weighted) offline dataset directly \citep{fujimoto2019off,wu2019behavior,kumar2019stabilizing,fujimoto2021minimalist,wang2020critic}; 2) {\em pessimism:} considering conservative value/Q-function by penalizing the values over OOD state-action pairs \citep{lyu2022mildly,kumar2020conservative,kostrikov2021offline,buckman2020importance}; 3) {\em in-sample:} learning without querying any OOD actions \citep{kostrikov2021offline,garg2023extreme}.

To achieve behavior regularization, a large portion of prior works choose to regularize the policy towards the behavior policy. As the behavior policy is unknown, they usually rely on access to an approximation of the behavior policy, or on access to a distance metric between any policy and the behavior policy \citep{kumar2019stabilizing,fakoor2021continuous}. However, either estimating the behavior policy as a multimodal conditional distribution or estimating the distance between any policy and $\pib$ is a difficult task in practice, requiring tailored and complex designs such as conditional variational
auto-encoders (CVAEs) \citep{fujimoto2019off,lyu2022mildly} or restricted distance metrics \citep{dadashi2021offline}. To circumvent this, a natural question is: {\em Can we steer a learning policy towards the behavior policy without requiring access to the said policy?}

To answer this, we propose to rely on the Q-function of the behavior policy $\pib$, denoted as $Q^{\pib}$, instead of relying directly on $\pib$. The Q-function $Q^{\pib}$ can not only play a similar role as $\pib$ in pushing the learning policy towards the behavior policy, but can also provide additional information such as the quality of the dataset/actions (e.g., larger $Q^{\pib}$ represents better quality). In addition, 
harnessing $Q^{\pib}$ in offline RL is promising and has several essential advantages compared to prior works relying on (an estimate of) the behavior policy $\pib$:

\begin{figure*}[bpt]
    \centering
    \includegraphics[width=1.0\textwidth]{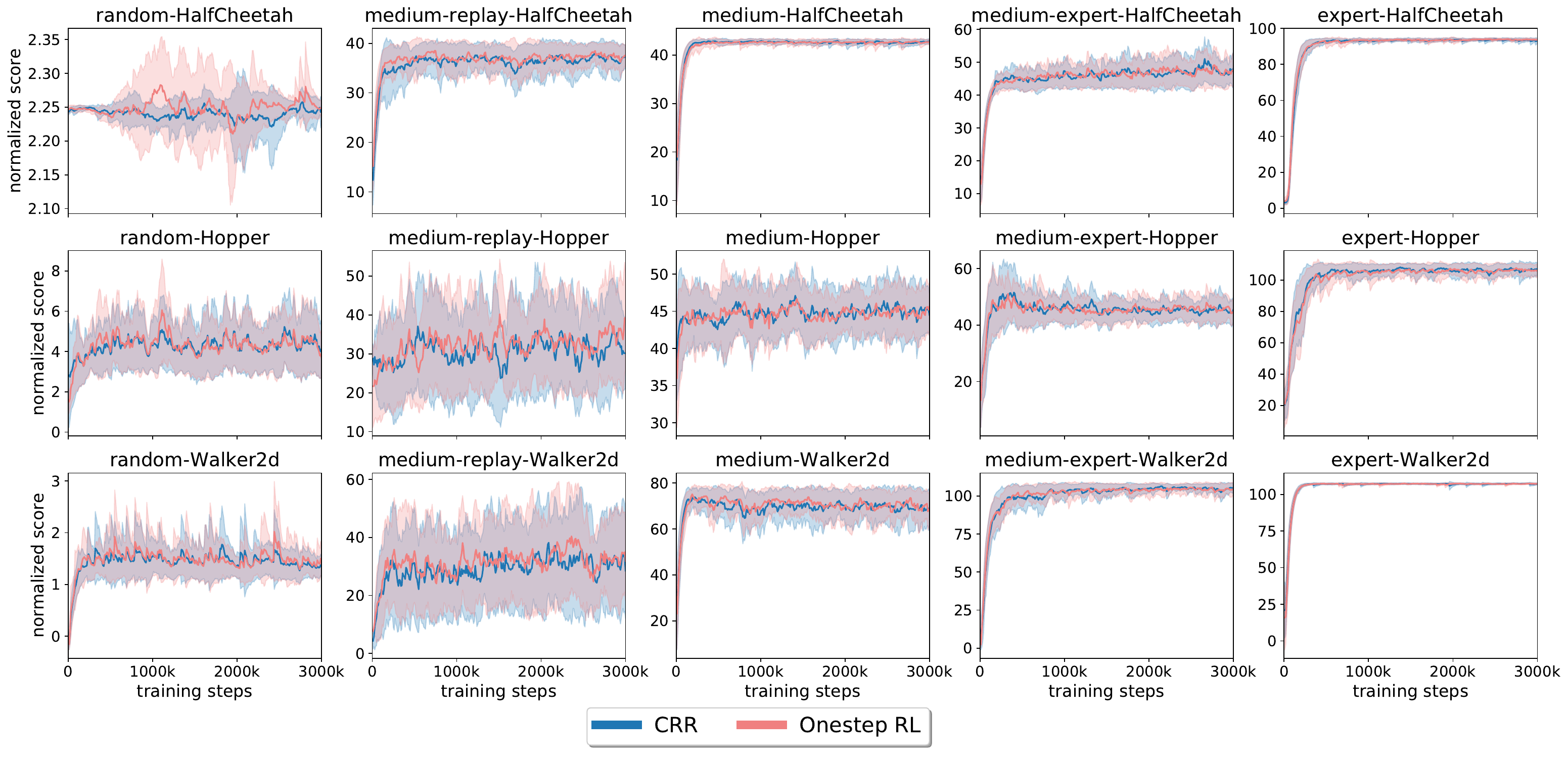}

    \caption{The comparisions between {\sf OnestepRL} \citep{brandfonbrener2021offline} and {\sf CRR} \citep{wang2020critic} over $15$ tasks of MuJoCo in D4RL \citep{fu2020d4rl}. {\sf CRR} and {\sf OnestepRL} use the same actor with different critics ({\sf OnestepRL} uses $\qsarsa$ while {\sf CRR} not). The similar performance implies that the actor plays a more important role compared to the critic $\qsarsa$ in {\sf OnestepRL}.}
    \label{fig:crr-1step-1row}

\end{figure*}

\begin{itemize}
 \item {\bf Estimating $Q^{\pib}$ is easier than $\pib$.} $Q^{\pib}$ can be estimated by $\qsarsa$ via minimizing a SARSA-style objective, while $\pib$ needs to be estimated as a multimodal conditional distribution, which is usually a more difficult problem.

\item {\bf Regularizing towards $Q^{\pib}$ is more straightforward than $\pib$.} The goal of imitating the behavior policy is usually to restrict extrapolation errors when evaluating the Q-function of the learning policy. Thus, it is more direct and efficient to regularize the Q-function of the learning policy towards $Q^{\pib}$ than to do so indirectly in the bootstrapping step via a regularization towards $\pib$.

\end{itemize}

Despite these advantages, there are few works exploring the use of $Q^{\pib}$ for offline RL regularization. Indeed, {\sf OnestepRL} \citep{brandfonbrener2021offline} is, to the best of our knowledge, the only work that promotes the usage of $Q^{\pib}$ in continuous action tasks. However, Fig.~\ref{fig:crr-1step-1row} (more details are specified in Sec.~\ref{eq:eval-qsarsa}) shows that although $\qsarsa$, the estimate of $Q^{\pib}$ using a SARSA-style objective, is one of the components in {\sf OnestepRL}, it may not be 
the critical one that contributes to its success and hence, more investigations are warranted.

We suggest that a proper use of $Q^{\pib}$ has the potential to improve the performance in offline RL. As a warm-up study, by adding a slight regularization towards $\qsarsa$ (the estimate of $Q^{\pib}$) in {\sf TD3} \citep{fujimoto2018addressing} without other modifications, we already observe a strong performance boost (see Fig.~\ref{fig:intro-motivate}).
Motivated by this potential, in this work, we propose to use $Q^{\pib}$ (estimated by $\qsarsa$) as a promising regularization methodology for offline RL. We first demonstrate that $\qsarsa$ is a reasonable estimate for $Q^{\pib}$ for {\em in-sample} state-action pairs and has controllable error over out-of-distribution regions. Equipped with these findings, we introduce two $\qsarsa$-regularized algorithms based on {\sf TD3-BC}, one of the state-of-the-art offline methods \citep{fujimoto2021minimalist}, and achieve strong performance on the D4RL benchmark \citep{fu2020d4rl}, notably better than {\sf TD3-BC} that we built on and {\sf OnestepRL}.

\section{Related works}
We shall discuss several lines of research on offline RL, with an emphasis on the most related works that make use of behavior regularization.

\paragraph{Offline RL.}
The major challenge of offline RL is the extrapolation error of policy evaluation induced by the distribution shift between the history dataset and the desired learning policy \citep{fujimoto2019off,kumar2019stabilizing}. As a result, a large portion of works follow the approximate dynamic programming framework (e.g., actor-critic) to balance learning beyond the behavior policy and handling the distribution shift. They usually resort to behavior regularization/constraints \citep{ghasemipour2021emaq,fujimoto2019off,kumar2019stabilizing,wu2019behavior,wang2020critic}, pessimistic value/Q-function estimation for out-of-distribution actions \citep{lyu2022mildly,kumar2020conservative,kostrikov2021offline,yu2021combo,rezaeifar2022offline}, or learning without querying any OOD actions \citep{kostrikov2021offline,garg2023extreme}. In addition, prior works also leverage other techniques such as sequential modeling \citep{chen2021decision,janner2021offline}, representation learning \citep{lee2020representation}, and diffusion models as policies \citep{wang2022diffusion}.

\paragraph{Offline RL with behavior regularization.}

In order to handle distribution shift, besides restricting the learning of policy almost to be in-distribution/in-sample \citep{wang2020critic,peng2019advantage} or leveraging imitation learning and working directly with the dataset \citep{fujimoto2021minimalist,chen2020bail}, many prior works regularize the policy to an explicit estimation of the behavior policy \citep{ghasemipour2021emaq,fujimoto2019off,lyu2022mildly,kumar2019stabilizing,yang2022regularized,zhang2022behavior} or characterizing the discrepancy between policies based on special distance or tailored approaches \citep{kumar2019stabilizing,fakoor2021continuous,dadashi2021offline}.

However, the estimation of either the behavior policy or the distance between policies is difficult in practice or requires special design. In this work, we propose the Q-function of the behavior policy as an essential and helpful component in offline RL, which can be estimated easily but has almost not been studied in the literature. Only a few works \citep{peng2019advantage,wang2020critic,gulcehre2021regularized} 
consider the usage of the Q-function of the behavior policy, but usually combined with some special optimization objective for the actor (e.g.,  advantage-weighted regression) which may take the most credit, or focusing on distinct tasks such as discrete action tasks \citep{gulcehre2021regularized}. To fill the gap, this work evaluates the reliability of estimating the Q-function of the behavior policy, and makes use of this estimate to design methods achieving competitive performance over the baselines.

\section{Problem formulation and notations}

\paragraph{Discounted infinite-horizon MDP.}
In this paper, we consider a discounted infinite-horizon Markov Decision Process (MDP) $\cM = \{\cS,\cA, P, r, \gamma\}$. Here, $\cS$ is the state space, $\cA$ is the action space, $P: \cS\times \cA \rightarrow \Delta(\cS)$ represents the dynamics of this MDP (i.e., $P(\cdot \mymid s,a)$ denote the transition probability from current state-action pair $(s,a)$ to the next state), $r: \cS \times \cA \rightarrow \mathbb{R}$ is the immediate reward function, $\gamma \in [0,1)$ is the discount factor. We denote a stationary policy, also called an action selection rule, as $\pi: \cS \rightarrow \Delta(\cA)$. 
The value function $V^\pi: \cS \rightarrow \mathbb{R}$ and Q-value function $Q^\pi: \cS\times \cA \rightarrow \mathbb{R}$ associated with policy $\pi$ are defined as 
$V^\pi(s) = \mathbb{E} \left[\sum_{t=0}^\infty \gamma^t r(s_t, a_t) \mid s_0 = s; \pi\right]$
and $Q^\pi(s,a) = \mathbb{E} \left[\sum_{t=0}^\infty \gamma^t r(s_t, a_t) \mid s_0 = s, a_0 = a; \pi\right]$, 
where the expectation is taken over the sample trajectory $\{(s_t, a_t)\}_{t\geq 0}$ generated following that $a_t \sim \pi(\cdot \mymid s_t)$ and $s_{t+1} \sim P(\cdot \mymid s_t, a_t)$ for all $t\geq 0$.

\paragraph{Offline/Batch dataset.}
We consider offline/batch RL, where we only have access to an offline dataset $\cD$ consisting of $N$ sample tuples $\{ s_i, a_i, r_i, s_i', a_i'\}_{i=1}^N$ generated following some behavior policy $\pib$ over the targeted environment. The goal of offline RL is to learn an optimal policy $\pi^\star$ given dataset $\cD$ which maximizes the long-term cumulative rewards,
$\pi^\star = \argmax_{\pi} \mathbb{E}_{s\sim \rho}\left[V^\pi(s)\right]$,
where $\rho$ is the initial state distribution.

\begin{figure*}[t]
    \centering
    \includegraphics[width=1.0\textwidth]{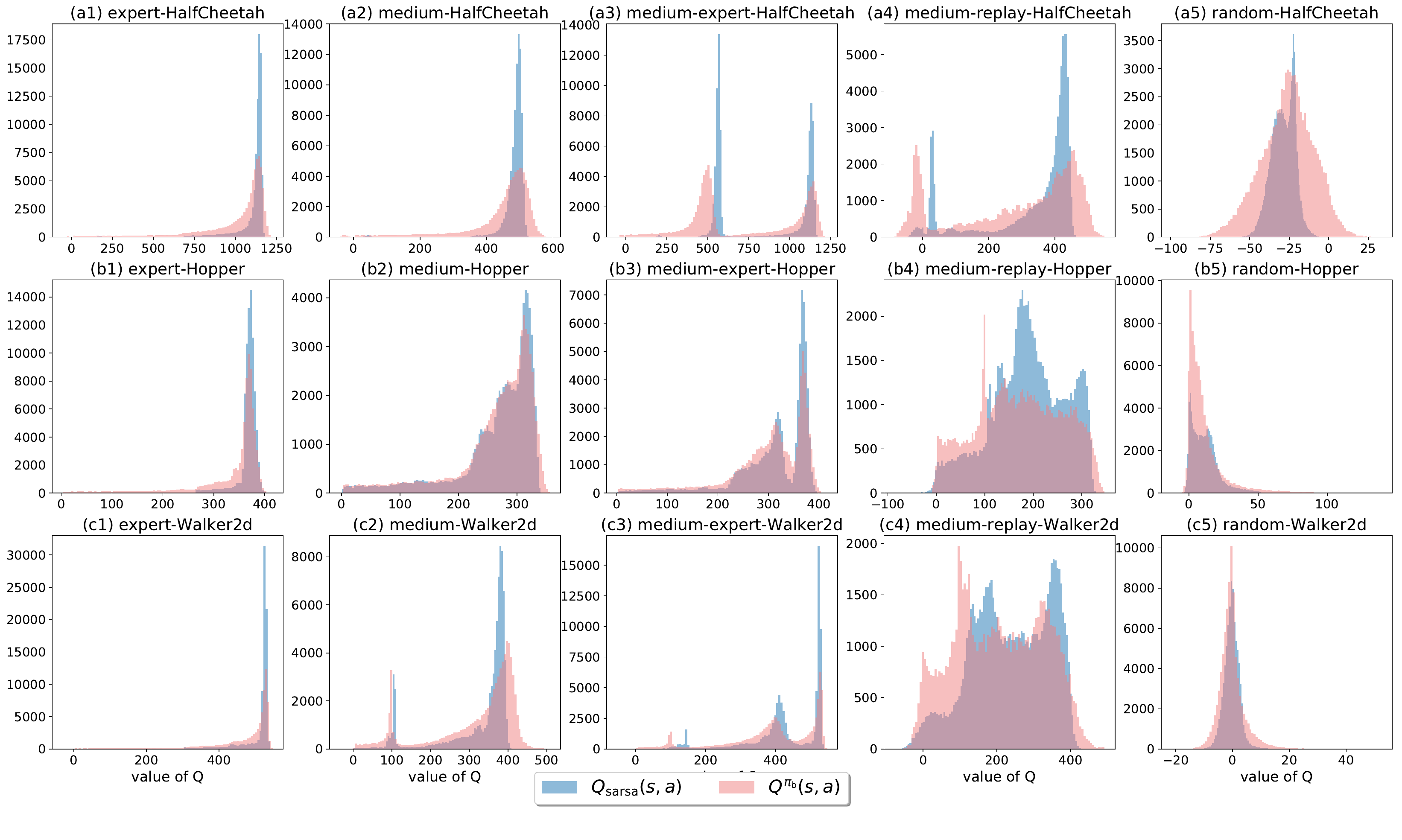}
    \vspace{-3mm}
    \caption{The demonstration of $\qsarsa$ and  $Q^{\pib}$ over $15$ tasks of MuJoCo in D4RL \citep{fu2020d4rl} with different tasks and diverse offline datasets. It shows that $\qsarsa$ estimate the value of $Q^{\pib}$ reasonably regarding that their distribution largely overlap with each other (red and blue histogorams).
    }
    \label{fig:qsarsa-property1}
\end{figure*}
\begin{figure*}[t]
    \centering
    \includegraphics[width=1.0\textwidth]{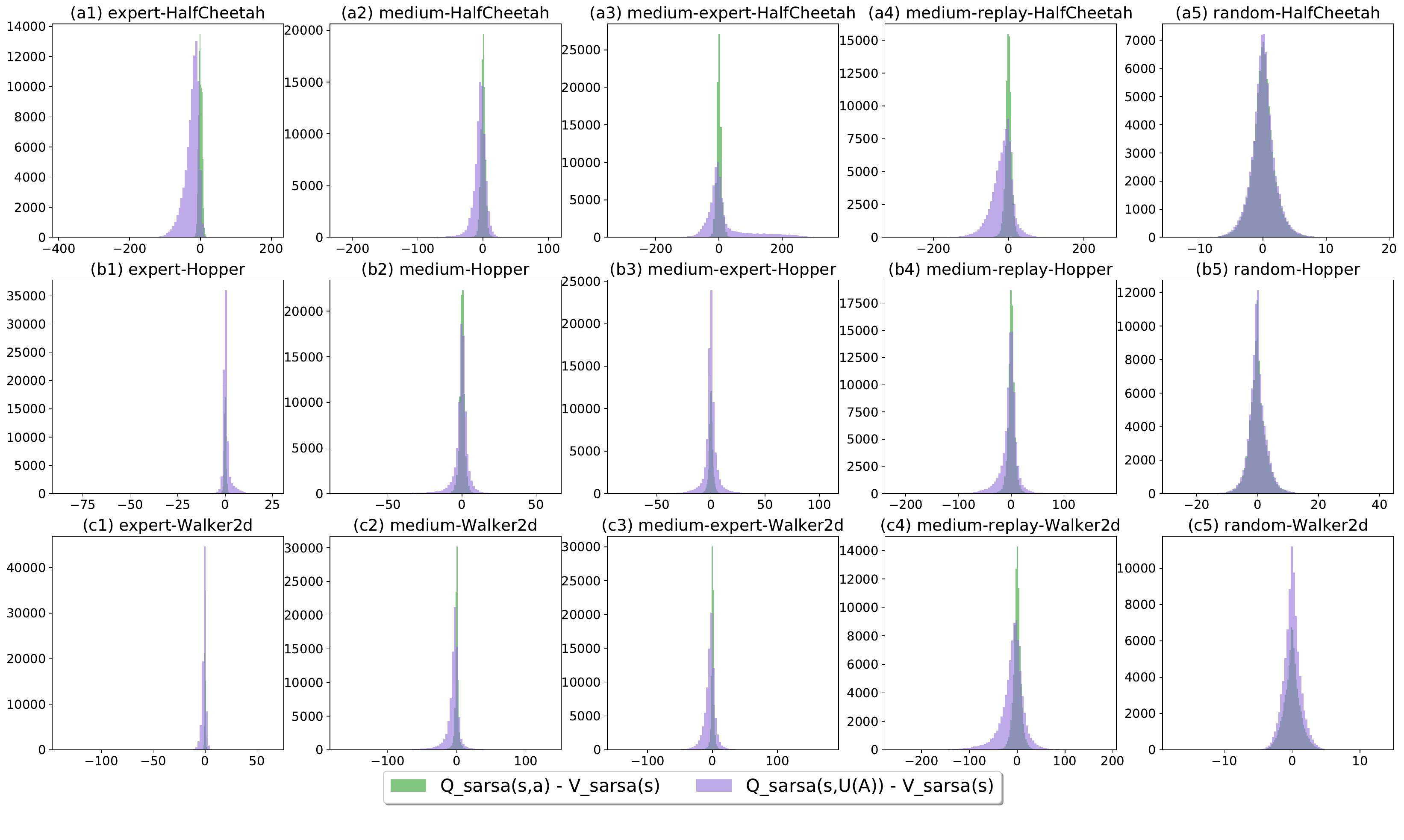}
    \vspace{-3mm}
    \caption{The illustration of the histograms of the advantage $\qsarsa(s,a) - \vsarsa(s)$ for {\em in-sample} state-action pairs $(s,a)\in\cD$ and also the advantage $\qsarsa(s,U(A)) - \vsarsa(s)$ for OOD state-action pairs $(s,U(A)) \notin \cD$ over $15$ tasks of MuJoCo in D4RL \citep{fu2020d4rl}. It indicates that $\qsarsa$ usually xihas no/small OOD error, since $\qsarsa$ over OOD state-action (purple histograms) won't exceed the value of in-distribution $\qsarsa$ (green histograms) too often or too much. }
    \label{fig:qsarsa-property2}
\end{figure*}

\section{Introduction and evaluation of $\qsarsa$}\label{sec:qsarsa}
In this work, we propose to use an estimate of the Q-function of the behavior policy ($Q^{\pib}$) for algorithm design. Since the investigation of exploiting $Q^{\pib}$ or its estimate in offline RL methods is quite limited in the literature, we first specify the form of the SARSA-style Q-estimate and then evaluate its reliability, characteristics, and possible benefits for prevalent offline algorithms.

\paragraph{The SARSA-style Q-estimate $\qsarsa$.}
We estimate $Q^{\pib}$ by the following SARSA-style optimization problem \citep{brandfonbrener2021offline} and denote the output as $\qsarsa$:
\begin{equation}
    \qsarsa \leftarrow \min_{\theta_s} \mathbb{E}_{\dc}[ (r(s,a) + \gamma Q_{\bar{\theta}_s}(s',a') - Q_{\theta_s}(s,a))^2 ],
    \label{eq:loss_sarsa}
\end{equation} 
where $\theta_s$ (resp.~$\overline{\theta}_s$) denotes the parameters of some neural network (resp.~target network, a lagging version of $\theta_s$), the expectation is w.r.t. $(s,a, r,s',a') \sim \cD$, and $\qsarsa \defn Q_{\theta_s}$. Note that $\qsarsa$ is estimated by {\em in-sample} learning, which only queries the samples inside the history dataset, without involving any OOD regions.

\subsection{Evaluation of $\qsarsa$}\label{eq:eval-qsarsa}

\paragraph{An ablation for {\sf OnestepRL}.}
First, to verify the benefit of using $\qsarsa$ directly as the critic in the actor-critic framework, proposed in {\sf OnestepRL} \citep{brandfonbrener2021offline}, we conduct an ablation study by replacing the critic $\qsarsa$ by the critic trained by vanilla temporal difference learning widely used in online RL (i.e., {\sf CRR} \citep{wang2020critic}) and show the comparisons in Fig.~\ref{fig:crr-1step-1row}. Fig.~\ref{fig:crr-1step-1row} show the ablation study results over $15$ MuJoCo tasks which include the main portion of the tasks considered in {\sf OnestepRL} \citep{brandfonbrener2021offline}. Note that to ensure a fair comparison, the actor component is the same for both methods --- advantage-weighted regression/exponentially weighted imitation \citep{brandfonbrener2021offline,wang2020critic}. Observing that the two methods perform almost the same in Fig.~\ref{fig:crr-1step-1row}, it indicates that the key to the success of {\sf OnestepRL} may be the objective function of the actor which restricts to {\em in-sample} learning but not $\qsarsa$, since replacing $\qsarsa$ does not impact the performance.

\paragraph{$\qsarsa$ estimates $Q^{\pib}$ reasonably well for $(s,a)\in\cD$.}
The goal of $\qsarsa$ is to estimate the ground truth $Q^{\pib}$. With this in mind, taking the MuJoCo tasks for instance, we provide the histograms of $\qsarsa(s,a)$ (in blue) and also  $Q^{\pib}(s,a)$ estimated by reward-to-go\footnote{Note that $Q^{\pib}$ is unknown. So we utilize the reward-to-go function \citep{janner2021offline} starting from any state-action pair $(s,a) \in\cD$ as $Q^{\pib}(s,a)$, i.e., $Q^{\pib}(s,a) \defn \sum_{t'=t}^T \gamma^{t'-t} r(s_{t'}, a_{t'})$ with $(s_t,a_t) = (s,a)$. The estimation can be filled by the trajectories in the entire dataset $\cD$ with simple Monte Carlo return estimate, the same as the estimation of the value function used by \citep{peng2019advantage}.} (in pink) in Fig.~\ref{fig:qsarsa-property1} for $(s,a) \in \mathcal{D}$ in different levels of history datasets. It shows that the distribution of the value of $\qsarsa(s,a)$ almost coincide with $Q^{\pib}(s,a)$, indicating that $\qsarsa$ estimates $Q^{\pib}$ accurately to some degree. The slight difference between $\qsarsa$ (blue histograms) and $Q^{\pib}$ (red histograms) might result from that $\qsarsa$ is estimating the Q-value of the behavior policy --- may be a mixing of a set of policies (such as in the {\em medium-replay} case), while each point of the red histograms from $Q^{\pib}(s,a)$ is actually the Q-value of one specific policy from the policy set but not the mixing policy (behavior policy).
\looseness=-1

In addition, the evaluation of $\qsarsa$ in Fig.~\ref{fig:qsarsa-property1} also provides useful intuition for the distribution of the dataset. For instance, the distribution of the {\em expert} datasets in Fig.~\ref{fig:qsarsa-property1}(a1)-(c1) concentrate around the region with larger value/Q-function value since they are generated by some expert policy, while the {\em medium-replay} datasets in Fig.~\ref{fig:qsarsa-property1}(a4)-(c4) have more diverse data over the entire space of the tasks (with more diverse Q-function values) due to they being sampled from some replay buffers generated by different policies.

\paragraph{$\qsarsa$ has controllable out-of-distribution (OOD) error.}
Before continuing, we first introduce a value function $\vsarsa: \cS \rightarrow \mathbb{R}$, where $\vsarsa(s) \defn \mathbb{E}_{a\in \pib(s)}\left[\qsarsa(s,a)\right]$ for all $s\in\cS$, learned by minimizing the following problem $\min_{V}\mathbb{E}_{(s,a)\in\cD}\left[|V(s) - \qsarsa(s,a)|^2\right]$. As mentioned in the introduction (Sec.~\ref{sec:intro}),  extrapolation error of the Q-value estimation, especially the overestimation over the out-of-distribution (OOD) state-action pairs is the essential challenge for offline RL problems. To evaluate if $\qsarsa$ overestimate the Q-value in OOD regions, in Fig.~\ref{fig:qsarsa-property2}, we demonstrate the deviations from $\qsarsa(s,a)$ to the expectation of $\qsarsa(s,a)$ ($\vsarsa(s)$) for state action pairs $(s,a)$ over {\em in-distribution} region $\cD$ (green histograms) and also OOD region ($(s, U(A))$ in purple histograms). In particular, $U(A)$ denotes the uniform distribution over the action space $\cA$.

Fig.~\ref{fig:qsarsa-property2} indicates that {\bf $\qsarsa(s,a)$ has controllable OOD error}. Specifically,  the relative $\qsarsa(s,a) -\vsarsa(s)$ value over OOD regions (purple histograms) generally has similar or smaller estimation value with comparisons to the maximum advantages over {\em in-sample} regions, i.e., $\max_{(s,a)\in\cD} \big(\qsarsa(s,a) -\vsarsa(s) \big)$ (green histograms). It implies that $\qsarsa$ 
usually has no/only slight overestimation over OOD regions and probably has some implicit pessimistic regularization effects.

\section{Offline RL with Q-sarsa (\our)}\label{sec:algo}
\newcommand{\topsepremove}{\aboverulesep = 1mm \belowrulesep = 1mm} \topsepremove

Inspired by the potential of $\qsarsa$ shown in Sec.~\ref{sec:qsarsa}, we propose a new algorithm called \our by taking advantage of $\qsarsa$ and designing regularizations based on it. To directly and clearly evaluate our changes around $\qsarsa$, \our is prototyped against one of the state-of-the-art offline methods --- {\sf TD3-BC} \citep{fujimoto2021minimalist} due to the simplicity of its algorithmic design. Specifically, among current offline state-of-the-art methods, we observe that {\sf TD3-BC} has the minimal adjustments --- by adding a {\sf BC} term in the actor --- compared to the existing off-policy method (used in online RL) \citep{fujimoto2018addressing}. As a result, it becomes much more straightforward and clear to evaluate the additional benefits, if any, indeed stem from our regularizations assisted by $\qsarsa$, rather than other components of the framework.

Before proceeding, we introduce several useful notations. We denote the parameter of two critic networks (resp.~target critic networks) as $\{\theta_i\}_{i\in\{0,1\}}$ (resp.~$\{\overline{\theta}_i\}_{i\in\{0,1\}}$). Similarly, $\phi$ (resp.~$\overline{\phi}$) represents the parameter of the policy (actor) network (resp.~ policy target network).

\subsection{Critic with $\qsarsa$}
Before introducing our method, we recall the optimization problem of the two critics in prior arts {\sf TD3}/{\sf TD3-BC} \citep{fujimoto2018addressing,fujimoto2021minimalist}, i.e., minimizing the temporal difference (TD) error: with $ y(s_j,a_j) = r(s_j, a_j) + \gamma \min_{i\in \{1,2\}} Q_{\overline{\theta}_i}(s_j', \widetilde{a}_j')$,
\begin{align}
    &\theta_i \leftarrow \min \underbrace{B^{-1}\sum_{ \cB}|y(s_j,a_j) - Q_{\theta_i}(s_j,a_j)|^2}_{\text{TD-error}}, \quad  i\in\{1,2\}, \label{eq:td-target}
\end{align} 
where $\cB = \{ s_j, a_j, r_j, s_j', a_j'\}_{i=1}^B$ is a batch of data sampled from $\cD$ with size $B$, $\widetilde{a}_j'$ is the perturbed action \citep{fujimoto2018addressing,fujimoto2021minimalist} chosen by current target policy given state $s_j'$, i.e., $\widetilde{a}_j' \approx \pi_{\overline{\phi} }(s_j')$.

With the above in mind, we propose the following critic object of \our by adding two natural regularization terms based on $\qsarsa$: for $i\in\{1,2\}$,
\begin{align}\label{eq:useful-critic}
     &\theta_i \leftarrow \min \text{TD-error}  + \lambda B^{-1} \sum_{ \cB} \Big\{ \underbrace{ \big[Q_{\theta_i}(s_j,a_j) - \qsarsa(s_j,a_j)\big]^2}_{\textcolor{red}{\mathsf{InD}\text{-}\mathsf{sarsa} } } \nonumber \\
    & \quad +  \underbrace{ w(s_j', \widetilde{a}_j' ) \cdot\big[  Q_{\theta_i}(s_j', \widetilde{a}_j') - Q_\text{sarsa}(s_j', \widetilde{a}_j')\big]^2}_{\textcolor{blue}{\mathsf{OOD}\text{-}\mathsf{sarsa}}} \Big\}, 
\end{align} 
where $\lambda$ is a constant factor and $w(\cdot)$ is a mask function to be specified shortly. 

Here, we choose the classic mean-squared error for both {\sf InD}-{\sf sarsa} and {\sf OOD}-{\sf sarsa} regularization terms. We introduce the role and intuition of the additional two regularizations separately.
\textbf{(i) {\sf InD}-{\sf sarsa}: in-distribution regularization.}
We highlight that this term regularizes the online learning Q-function (critic) $Q_{\theta_i}$ towards $Q_{\text{sarsa}}$ only counting on {\em in-sample} state-action pairs $(s_j,a_j)$ from the offline dataset (i.e., $(s_j,a_j)\in \mathcal{D}$). Recalling that for all $(s_j,a_j)\in \mathcal{D}$, $Q_{\text{sarsa}}(s_j,a_j)\approx Q^{\pib}(s_j,a_j)$ (see Fig.~\ref{fig:qsarsa-property1}) is a reasonable target, this {\em in-sample} regularization plays the role of pushing $Q_{\theta_i}$ to $Q^{\pib}$ to avoid the overestimation of $Q_{\theta_i}$. Note that this regularization is unlikely to bring in OOD errors for $Q_{\theta_i}$ since it 
 only acts on state-action pairs inside the dataset ($(s_j,a_j)\in\cD$). 
\textbf{(ii) {\sf OOD}-{\sf sarsa}: out-of-distribution regularization.}
In contrast to {\sf InD}-{\sf sarsa}, this term pushes the Q-function $Q_{\theta_i}$ towards $Q_{\text{sarsa}}$ by acting on OOD regions (i.e., $(s_j', \widetilde{a}_j')$ perhaps not in $\cD$). It is used to restrict/reduce the overestimation error of $Q_{\theta_i}$ over the OOD regions in order to circumvent the extrapolation error challenges.
\looseness=-1

Specifically, recall that the bootstrapping term $Q_{\overline{\theta}_i}(s_j', \widetilde{a}_j')$ in \eqref{eq:td-target} plays an essential role in estimating the Q-value, which has potentially catastrophic extrapolation error.
 The reason is that the state-action pair $(s_j', \widetilde{a}_j')$ may appear scarcely or be unseen in the offline dataset, yielding large OOD errors since $Q_{\theta_i}(s_j', \widetilde{a}_j')$ and/or $Q_{\overline{\theta}_i}(s_j', \widetilde{a}_j')$ may not be sufficiently optimized  during training. To address this, {\sf OOD}-{\sf sarsa} directly regularizes $Q_{\theta_i}(s_j', \widetilde{a}_j')$ towards a more stable Q-function $\qsarsa(s_j',\widetilde{a}_j')$, which generally does not incur large OOD errors (see Sec.~\ref{sec:qsarsa}), to restrict the overestimation error of the bootstrapping term. 
Last but not least, we specify the choice of $ w(\cdot,\cdot)$, which is a hard mask to prevent the regularization term from being contaminated by extra OOD errors. Specifically, we remove {\em bad state-action pairs} in case $\qsarsa(s_j', \widetilde{a}_j')$ has relatively large OOD error: $w(\cdot,\cdot)$ is set as
    \begin{align}
    w(s_j',\widetilde{a}_j') &=\mathbb{1}\Big(\qsarsa(s_j',\widetilde{a}_j') > \vsarsa(s_j') - |\qsarsa(s_j',a_j') - \vsarsa(s_j')| \Big). \label{eq:beta-critic}
\end{align} 

\begin{table*}[ht]

\caption{The normalized score (see Appendix~\ref{sec:metric}) of the final 10 evaluations over $5$ random seeds. The highest performing scores are highlighted as pink and $\pm$ captures the standard deviation over seeds. For all the baselines, we re-run the implementations in the Acme framework \citep{hoffman2020acme} to keep an identical evaluation process for all methods. It shows that our two methods \our and \ourr achieve better performance over the $15$ tasks of MuJoCo benchmarks in D4RL \citep{fu2020d4rl}.
	} 
	\label{tab:exp-result} 

\begin{center}
\resizebox{\textwidth}{!}{
\begin{tabular}{cc|ccccccc}
\hline 
\toprule
  & & \multirow{2}{*}{\bf BC} & \multirow{2}{*}{\bf CRR} & \multirow{2}{*}{\bf OnestepRL} & \multirow{2}{*}{\bf CQL} & \multirow{2}{*}{\bf TD3-BC} & {\bf Qsarsa-AC}  & {\bf Qsarsa-AC2} \tabularnewline
  & &&&&&& (Ours) & (Ours) \tabularnewline
\hline 
\toprule 
\multirow{3}{*}{{\rotatebox[origin=c]{90}{Random}}}  & $\mathsf{HalfCheetah}$  &$2.3\pm0.0$
&$2.2\pm0.1$ &$2.3\pm0.0$ &$25.6\pm1.8$ &$6.0\pm1.6$ & \colorbox{tacell!30}{$30.2$}$\pm1.3$
& \colorbox{tacell!30}{$30.1$}$\pm1.3$ \tabularnewline
 & $\mathsf{Hopper}$  &$4.1\pm1.9$
&$3.9\pm1.3$
&$3.6\pm1.0$
&$8.7\pm1.3$ 
& \colorbox{tacell!30}{$12.7$}$\pm9.3$
&$7.5\pm0.9$
&$8.1\pm1.7$\tabularnewline
  &$\mathsf{Walker2d}$ &$1.3\pm0.2$
&$1.4\pm0.2$
&$1.6\pm0.3$
&$0.3\pm0.5$
& \colorbox{tacell!30}{$5.4$}$\pm6.3$
& \colorbox{tacell!30}{$5.9$}$\pm8.1$
&$4.5\pm4.9$ \tabularnewline
\hline
\multirow{3}{*}{\rotatebox[origin=c]{90}{\parbox[c]{1.2cm}{\centering Medium Replay}}}  & $\mathsf{HalfCheetah}$  &$34.4\pm3.1$
&$37.3\pm2.1$
&$38.1\pm2.1$
&$-2.4\pm1.9$
&$42.5\pm0.7$
&$42.2\pm0.7$
& \colorbox{tacell!30}{$54.4$} $\pm1.8$ \tabularnewline
  & $\mathsf{Hopper}$ &$30.0\pm9.5$
&$29.9\pm7.8$
&$43.2\pm9.2$
&\colorbox{tacell!30}{$93.0$}$\pm2.4$
&$39.3\pm8.4$
& \colorbox{tacell!30}{$92.6$}$\pm10.2$
&$68.1\pm19.7$  \tabularnewline
 & $\mathsf{Walker2d}$ &$21.3\pm17.1$
&$21.1\pm12.2$
&$28.6\pm12.3$
&$13.9\pm13.0$
&$67.6\pm6.4$
& \colorbox{tacell!30}{$83.7$}$\pm9.0$
&$80.0\pm11.4$  \tabularnewline
\hline
\multirow{3}{*}{{\rotatebox[origin=c]{90}{Medium}}} & $\mathsf{HalfCheetah}$ &$42.5\pm2.0$
&$42.7\pm0.7$
&$42.8\pm0.7$
&$48.7\pm0.4$
&$45.4\pm0.4$
&$44.1\pm0.5$
& \colorbox{tacell!30}{$56.4$}$\pm0.8$  \tabularnewline
 & $\mathsf{Hopper}$ &$41.1\pm10.2$
&$43.3\pm2.0$
&$45.5\pm3.2$
&$55.6\pm6.2$
&$46.0\pm3.3$
& \colorbox{tacell!30}{$78.6$}$\pm15.7$
&$66.6\pm24.3$  \tabularnewline
  & $\mathsf{Walker2d}$ &$67.4\pm12.6$
&$65.6\pm11.5$
&$70.7\pm10.1$
&$76.6\pm6.6$
& \colorbox{tacell!30}{$82.5$}$\pm0.9$
&$79.0\pm3.1$
& \colorbox{tacell!30}{$83.3$}$\pm3.5$
\tabularnewline
\hline
\multirow{3}{*}{\rotatebox[origin=c]{90}{\parbox[c]{1.2cm}{\centering Medium Expert}}}  & $\mathsf{HalfCheetah}$ &$50.2\pm6.9$
&$49.6\pm8.2$
&$47.9\pm6.3$
&$68.4\pm16.0$
&$91.8\pm1.4$
& \colorbox{tacell!30}{$91.3$}$\pm0.9$
&$90.9\pm1.0$  \tabularnewline
 & $\mathsf{Hopper}$ &$48.7\pm6.8$
&$47.5\pm3.2$
&$45.6\pm4.3$
& \colorbox{tacell!30}{$92.7$}$\pm16.1$
&$83.6\pm17.6$
&$56.9\pm21.3$
&$73.6\pm25.5$  \tabularnewline
  & $\mathsf{Walker2d}$  &$98.0\pm14.3$
&$105.6\pm2.6$
&$105.5\pm3.4$
&$106.2\pm1.4$
&$106.4\pm5.2$
&$105.2\pm6.1$
& \colorbox{tacell!30}{$107.5$}$\pm1.2$ \tabularnewline
\hline
\multirow{3}{*}{{\rotatebox[origin=c]{90}{Expert}}} & $\mathsf{HalfCheetah}$ &$91.3\pm2.4$
&$93.9\pm0.8$
&$93.8\pm1.2$
&$72.6\pm37.7$
& \colorbox{tacell!30}{$94.6$}$\pm0.4$
&$93.8\pm1.0$
&$93.7\pm1.0$  \tabularnewline
 & $\mathsf{Hopper}$ &$105.3\pm6.6$
& \colorbox{tacell!30}{$107.3$}$\pm4.9$
&$105.9\pm5.9$
&$75.6\pm26.9$
&$104.2\pm5.5$
&$95.0\pm21.1$
&$97.8\pm17.0$  \tabularnewline
  & $\mathsf{Walker2d}$ &$107.5\pm0.3$
&$107.2\pm0.6$
&$107.3\pm0.3$
&$106.0\pm5.0$
& \colorbox{tacell!30}{$108.1$}$\pm0.3$
&$107.0\pm0.6$
&$107.0\pm0.8$  \tabularnewline
\hline
\toprule
& Total &$745.3\pm94.1$
&$758.5\pm58.1$
&$782.4\pm60.3$
&$841.5\pm137.2$
&$936.2\pm67.8$
& \colorbox{tacell!30}{$1013.2$}$\pm100.3$
& \colorbox{tacell!30}{$1021.8$}$\pm116.0$
\tabularnewline
\hline
\toprule

\end{tabular}
}
\end{center}

\end{table*}

\subsection{Actor with $\qsarsa$}

Recall the ideal optimization problem for the learning policy (actor) in {\sf TD3-BC}~\citep{fujimoto2021minimalist}:
$\max_\pi \mathbb{E}_{\mathcal{D}}[ \frac{\alpha}{\mathbb{E}_{\mathcal{D}} [Q(s,a)]} \cdot Q(s,\pi(s)) - (\pi(s)-a)^2 ]$, 
which directly adds a behavior cloning ({\sf BC}) regularization towards the distribution of the dataset with a universal weight $\alpha$ for any offline dataset and any state-action pair. In this work, armed with $\qsarsa$, we propose the following optimization problem referring to $\qsarsa$-determined point-wise weights $f(\qsarsa,s,a)$ for {\sf BC} instead of the fixed universal $\alpha$:
\begin{align}
    \max_\pi \mathbb{E}_{\mathcal{D}}\left[\frac{ Q(s,\pi(s))}{\mathbb{E}_{\mathcal{D}} [Q(s,a)] }    - \textcolor{red}{f(\qsarsa, s, a)}  (\pi(s)-a)^2 \right], \label{eq:our-actor}
\end{align}
where $f(\qsarsa,s,a) \defn p_{s,a}(\qsarsa)/ g(\qsarsa)$ is constructed by two terms given below. {\bf (i) Global weight $g(\qsarsa)$ for BC.} $g(\qsarsa)$ serves as a global quantification of the dataset quality and determines the foundation of the weights on BC (keeping the same for all $(s,a)\in\cD$). Intuitively, when the dataset is generated by some high-quality policies (e.g., the {\em expert} dataset), we are supposed to imitate the policy and put more weights on the BC regularization (bigger $1/g(\qsarsa)$), otherwise smaller weights on BC. Towards this, we use $\qsarsa^{\mathsf{mean}} \defn \big|\mathbb{E}_{ \mathcal{D}}[\qsarsa(s,a)]\big|$ as a global quantification for the behavior policy, which leads to
    \begin{align}
    g(\qsarsa) = \text{Clip}\left[\alpha  \exp\left(\frac{\tau_1}{\qsarsa^{\mathsf{mean}}}\right),(1,10^{6})\right]. \label{eq:g-qsarsa}
\end{align} 
Here, the clipping function $\text{Clip}(\cdot)$ is just to normalize $g(\qsarsa)$ between $1$ and a huge number (e.g., $10^6$) to forbid it from being too big, which can cause numerical problems.
{\bf (ii) Point-wise weight $p_{s,a}(\qsarsa)$ for BC.}
$p_{s,a}(\qsarsa)\in[0,1]$ is typically a point-wise normalized weight for different $(s,a)$ pairs, formed as
\begin{align}
    p_{s,a}(\qsarsa) =\max\left( \exp\Big(\frac{\tau_2 \left(\qsarsa(s,a) -\vsarsa(s) \right)}{ \big|\qsarsa(s,a) \big|}\Big), 1 \right).
\end{align} 
In particular, $p_{s,a}(\qsarsa)$ puts larger weights on {\sf BC} for {\em high-quality} state-action pair $(s,a)$ which deserves the learning policy to visit, otherwise reduces the weights for {\sf BC} regularization. The quality of the state-action pairs is determined by the advantage $\qsarsa(s,a) - \vsarsa(s)$ normalized by $\qsarsa(s,a)$.

\subsection{A variant \ourr}
Given that the global weight $g(\qsarsa)$ aims to evaluate the quality of the given dataset over some specific task, it is supposed to depend on the relative value of $Q^{\pib}$ w.r.t. the optimal Q-function $Q^\star$ over this task.
However, $g(\qsarsa)$ in \eqref{eq:g-qsarsa} is calculated by the absolute value of $\qsarsa \approx Q^{\pib}$ without considering that $Q^\star$ may vary in different tasks (for example, $Q^\star$ of hopper or walker2d are different). So supposing that we have access to $\max_{s,a} Q^\star(s,a)$ for different tasks (can be approximated by the maximum of the reward function), we propose \ourr as a variant of \our which only has a different form of $g(\qsarsa)$ compared to \our as follows:
\begin{equation}
 g(\qsarsa)  = \text{Clip}\left[\alpha  \exp\left(\frac{\tau_3 \max_{s,a}   Q^\star(s,a)}{\qsarsa^{\mathsf{mean}}}\right),(1,10^{6})\right],
 \end{equation}
where $\max_{s,a} Q^\star(s,a)$ is estimated by $ \frac{\max_{s,a} r_{\text{expert}}(s,a) }{ (1-\gamma)}$ with $r_{\text{expert}}(s,a)$, the reward data from the {\em expert} dataset.

\begin{figure*}[ht]
    \centering
    \includegraphics[width=1.0\textwidth]{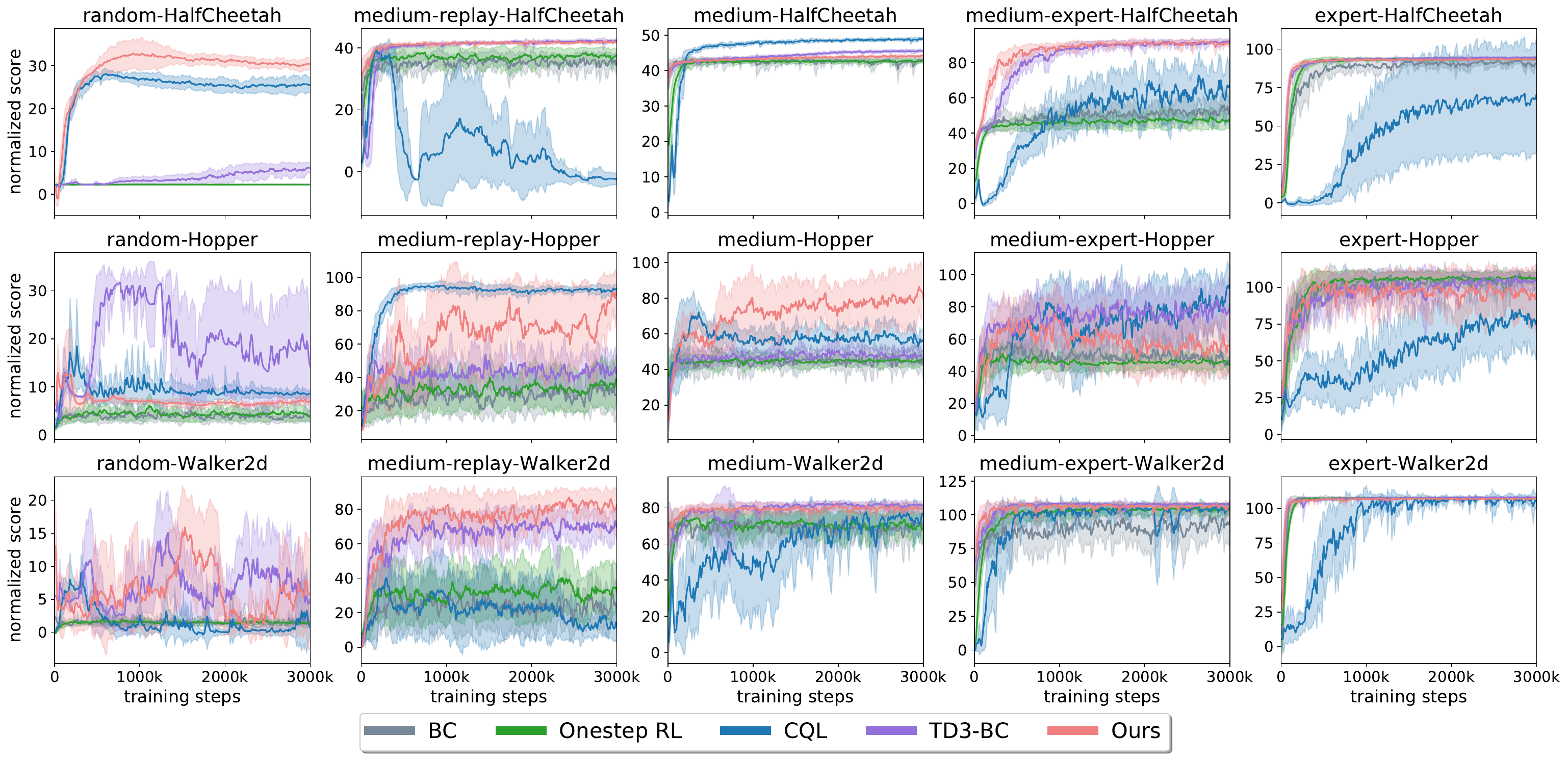}
    \vspace{-4mm}

    \caption{Normalized scores (see Appendix~\ref{sec:metric}) of the evaluations during the training process (5 seeds).}
    \label{fig:exp-result}

\end{figure*}

\begin{figure*}[ht]
    \centering
    \includegraphics[width=1.0\textwidth]{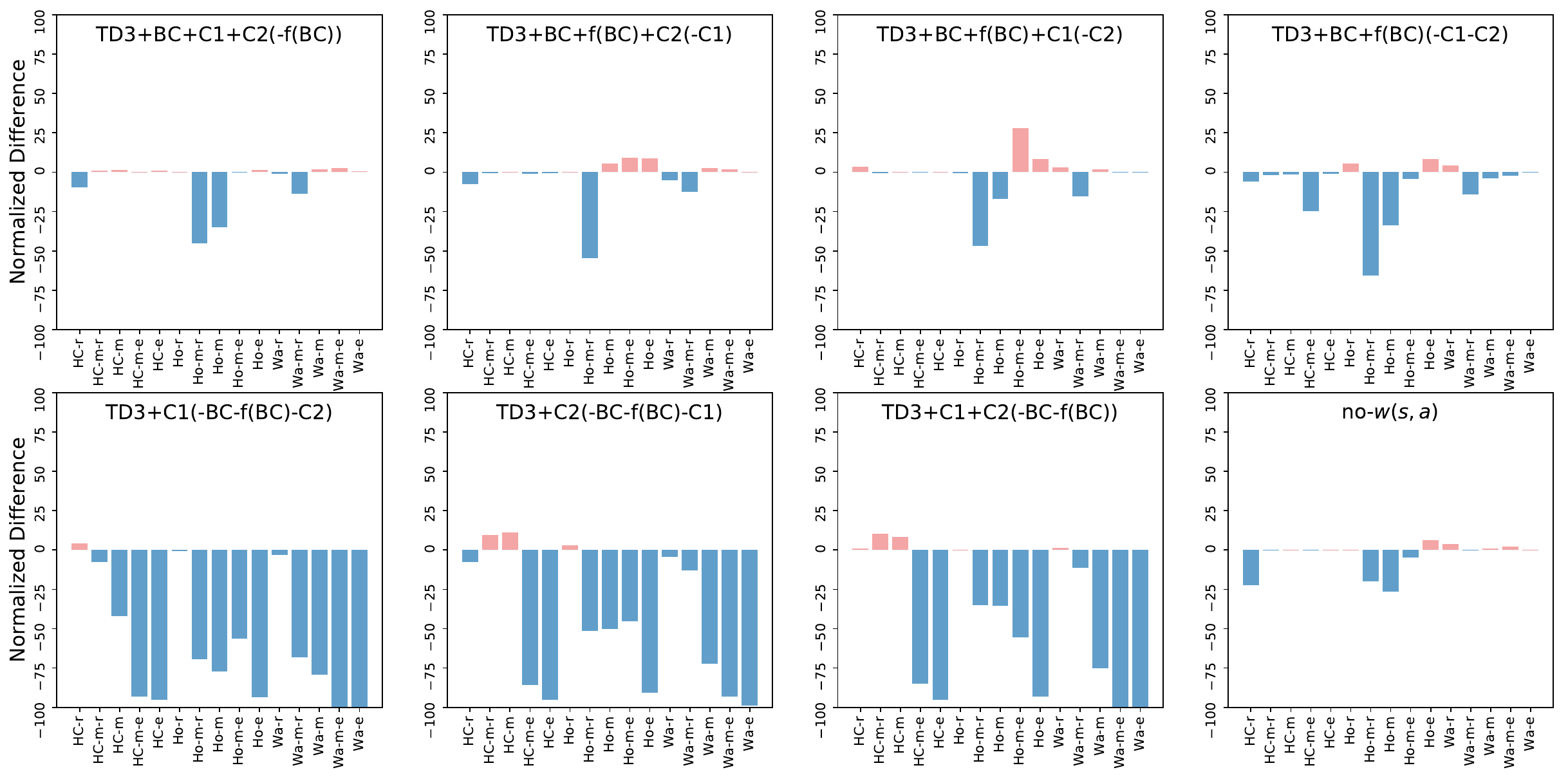}
   
    \caption{The ablation study of \our (denoted as {\sf TD3 + BC + f(BC) + C1 + C2}), where {\sf f(BC)} means the weight $f(\qsarsa,s,a)$ for BC in \eqref{eq:our-actor}, and {\sf C1} (resp.~{\sf C2}) denotes the $1^\text{st}$ (resp. $2^\text{nd}$) regularization term designed for the critic in \eqref{eq:useful-critic}. Here, the {\em Normalized Difference} is calculated as the difference of the normalized score (see Appendix~\ref{sec:metric}) between one ablation and the full algorithm \our (the mean score of $10$ evaluations over $5$ seeds).
   .
    }
    \label{fig:ablation}

\end{figure*}

\section{Experimental evaluation}\label{sec:experiments}

We first introduce the evaluation of our methods with comparisons to state-of-the-art baselines over the D4RL MuJoCo benchmarks \citep{fu2020d4rl}, followed by ablation experiments to offer a more detailed analysis for the components based on $\qsarsa$.

\subsection{Settings and baselines}

\paragraph{Experimental settings.}
To evaluate the performance of the proposed methods, we conduct experiments on the D4RL benchmark of OpenAI gym MuJoCo tasks \citep{fu2020d4rl}, constructed with various domains and dataset quality. Specifically, we conduct experiments on the recently released `-v2' version for MuJoCo in D4RL consisting of $5$ different levels of offline datasets ({\em random}, {\em medium-replay}, {\em medium}, {\em medium-expert}, and {\em expert}) over $3$ different environments, in total $15$ tasks. All the baselines and our methods are trained for 3M steps.

\paragraph{Baselines.}
Besides behavior cloning ({\sf BC}), we compare our performance to several state-of-the-art offline methods, namely {\sf CRR} \citep{wang2020critic}, {\sf OnestepRL} \citep{brandfonbrener2021offline}, {\sf CQL} \citep{kumar2020conservative}, and {\sf TD3-BC} \citep{fujimoto2021minimalist}. We highlight that {\sf OnestepRL} and {\sf TD3-BC} are the most related baselines: 1) {\sf OnestepRL} is the only prior work that exploit $\qsarsa$ in offline methods by directly setting $\qsarsa$ as the critic, whereas we adopt slight regularizations with the aid of $\qsarsa$; 2) our methods are designed with {\sf TD3-BC} as the prototype with additional $\qsarsa$-assisted regularizations. For all the baselines, we use the implementations from the Acme framework \citep{hoffman2020acme} which maintains the designing details of the original papers with tuned hyper-parameters.

\paragraph{Implementations and hyperparameters.}
Recall that the core of the proposed methods is $\qsarsa$ --- an estimate of $Q^{\pib}$ --- can be learned from \eqref{eq:loss_sarsa}. Hence, the training of our methods is divided into two phases: 1) learning $\qsarsa$ by solving \eqref{eq:loss_sarsa} for 1M steps; 2) following the learning process of {\sf TD3-BC} with a fixed $\qsarsa$ for 3M training steps.

The hyperparameters of our methods defined in Sec.~\ref{sec:algo} are tuned in a small finite set using $5$ random seeds that are different from those used in the final results reported in Table~\ref{tab:exp-result}. In particular, we use $\lambda= 0.03$ for the critic and $\alpha = 10^{-4}$, $\tau_1 =3000, \tau_2 = 4, \tau_3 = 4.5$ for the actor. We remark that $\tau_1$ is chosen according to the range of the value of $\qsarsa$ as $\tau_1 \approx 2 \cdot  \max_{s,a}\qsarsa(s,a)$ in MuJoCo. For the critic, we tune $\lambda$ in \eqref{eq:useful-critic} within the set $\{0.01, 0.03, 0.05\}$. For the actor, $\tau_2$ is tuned inside $\{0.4, 2,4\}$, and $\tau_1$ and $\tau_3$ are tuned across $\{2500, 3000, 3500\}$ and $\{4,4.5,5\}$ with a fixed $\alpha = 10^{-4}$.

\subsection{Main results}

\paragraph{Evaluation of $\qsarsa$.}
We evaluate our two methods \our and \ourr with comparison to the baselines over the $15$ tasks using $5$ random seeds; the results are reported in Table~\ref{tab:exp-result}.
There are three key observations. {\bf (i) $\qsarsa$ brings benefits to existing offline methods.} Recall that our proposed methods are built on the framework of {\sf TD3-BC}. The last 3 columns of Table~\ref{tab:exp-result} illustrate the comparisons between our two methods and {\sf TD3-BC}. It indicates that the proposed \our and \ourr methods, which take advantage of the information of $\qsarsa$ (see Sec.~\ref{sec:algo}), can bring additional benefits to the existing offline framework {\sf TD3-BC} and 
outperform all the conducted baselines. It is also promising to integrate $\qsarsa$ to other offline methods using approximate dynamic programming framework such as {\sf SAC} \citep{haarnoja2018soft} and {\sf IQL} \citep{kostrikov2021offline2}. We also 
show that our methods has competitive results against some additional state-of-the-art methods (such as {\sf IQL} and {\sf DT}) in Table~\ref{tab:exp-result2} since these baselines are not closely related.
{\bf (ii) Directly setting $\qsarsa$ as the critic has no benefits.}
    To evaluate whether $\qsarsa$ play an important role in the success of {\sf OnestepRL} \citep{brandfonbrener2021offline} more systematically, we also involve {\sf CRR} as an ablation baseline to {\sf OnestepRL}. As introduced in Sec.~\ref{sec:qsarsa}, we use variants of {\sf OnestepRL} and {\sf CRR} with the same actor loss (exponentially weighted imitation \citep{brandfonbrener2021offline}), so that the only difference between {\sf OnestepRL} and {\sf CRR} is that {\sf OnestepRL} uses $\qsarsa$ as the critic, while {\sf CRR} uses the critic learned by TD error (usually used in online RL).
    The similar scores in Table.~\ref{tab:exp-result} and also the normalized training scores in Fig.~\ref{fig:crr-1step-1row} show that there is almost no difference between {\sf CRR} and {\sf OnestepRL}, indicating that the success of {\sf OnestepRL} may be attributed to the actor loss.
{\bf (iii) Regularization strength for $\qsarsa$.} Based on Table~\ref{tab:exp-result}, it is noted that our methods achieve better performance with very small weights on the critic regularization terms ($\lambda= 0.03$ in \eqref{eq:useful-critic}). $\qsarsa$ typically doesn't influence the actor widely since it only determines the weights of the {\sf BC} term. This observation implies that some small regularizations based on the Q-function of $\pib$ ($\qsarsa$) may already achieve the goal of addressing the extrapolation and OOD error in offline RL.

\paragraph{Ablation study.}

We also perform an ablation study for the components of our method \our, illustrated in Fig.~\ref{fig:ablation}. Recall that the main components of \our include $f(\qsarsa,s,a)$ (denoted as {\sf f(BC)}) for the actor and two regularizations for the critic --- {\sf InD}-{\sf sarsa} (represented as {\sf C1}) and {\sf OOD}-{\sf sarsa} (represented as {\sf C2}).
In Fig.~\ref{fig:ablation}, the first row shows that if we keep {\sf TD3-BC} as the baseline, removing any of the three components based on $\qsarsa$, especially removing the two critic regularizations together, leads to evident dropping of the performance. It indicates that the information of $\qsarsa$ inside the components brings significant benefits to the existing  {\sf TD3-BC} framework. We further evaluate the ablations of removing {\sf BC} in the actor (as well as {\sf f(BC)}), which lead to dramatic dropping of the performance, shown in the second row of Fig.~\ref{fig:ablation}. The performance drop is reasonable since the weights for the critic regularizations ({\sf C1} and {\sf C2}) designed in \eqref{eq:useful-critic} are tuned based on {\sf TD3-BC} with {\sf BC} in hand. Finally, the ablation of removing the mask $w(\cdot)$ inside the {\sf OOD}-{\sf sarsa} regularization in \eqref{eq:useful-critic} implies that OOD error of $\qsarsa$ may happen and 
harm the performance without $w(\cdot)$, but not often.

\section{Conclusion}
We propose to integrate a SARSA-style estimate of the Q-function of the behavior policy into offline RL for better performance. Given the limited use of the Q-function of the behavior policy in the current literature, we first evaluate the SARSA-style Q-estimate to establish its reliability in estimating the Q-function and potential to restrict OOD errors. We then propose two methods by taking advantage of the SARSA-style Q-estimate based on {\sf TD3-BC}, one of the offline state-of-the-art methods. Our proposed methods achieve strong performance in D4RL MuJoCo benchmarks and outperform the baselines. It is our hope to  inspire future works that exploit the benefit of Q/value-functions of the behavior policy in more offline methods, such as using different regularization loss functions beyond $\ell_2$, combining it with other regularization techniques, and in different schemes in the approximate dynamic programming framework, even sequential modeling.

\section*{Acknowledgement} Part of this work was completed when L. Shi was an intern at Google Research, Brain Team. The work of L. Shi and Y. Chi is supported in part by the grants NSF CCF-2106778 and CNS-2148212. L. Shi is also gratefully supported by the Leo Finzi Memorial Fellowship, Wei Shen and Xuehong Zhang Presidential Fellowship, and
Liang Ji-Dian Graduate Fellowship at Carnegie Mellon University. The authors would like to thank Alexis Jacq for reviewing an early version of the paper. The authors would like to thank the anonymous reviewers for valuable feedback and suggestions. We would also like to thank the Python and RL community for useful tools that are widely used in this work, including Acme \citep{hoffman2020acme}, Numpy \citep{harris2020array}, and JAX \citep{bradbury2021jax}.

\bibliographystyle{apalike}
\bibliography{q-sarsa}

\appendix

\section{Appendix}
\subsection{The comparisons to additional baselines.}\label{sec:extra-exp}

Here, we provide the performance comparisons to two additional strong baselines {\sf DT} \cite{chen2021decision} and {\sf IQL} \cite{kostrikov2021offline2}.  {\sf IQL} is a well-known strong baseline using the same approximate dynamic programming framework as our methods, while {\sf DT} resort to a different framework --- sequential modeling. In Table \ref{tab:exp-result2}, we directly report the scores in the original paper of {\sf DT} \cite{chen2021decision} and the reported scores for {\sf IQL} in a prior work \cite{lyu2022mildly}. Table \ref{tab:exp-result2} shows that our methods can not only bring significant benefits for some existing offline methods (such as {\sf TD3-BC}) but also achieve competitive performance with comparisons to those strong offline methods.
\begin{table}[H]

\begin{center}
	\caption{This table displays the comparisons with two additional offline state-of-the-arts {\sf DT} \cite{chen2021decision} and {\sf IQL} \cite{kostrikov2021offline2} over the $15$ tasks of MuJoCo benchmarks in D4RL \cite{fu2020d4rl}.
	} 
\resizebox{0.85\textwidth}{!}{
\begin{tabular}{cc|cccc}
\hline 
\toprule
   & & \multirow{2}{*}{\bf DT} & \multirow{2}{*}{\bf IQL} & {\bf \our}  & {\bf \ourr} \tabularnewline
   & && & (Ours) & (Ours) \tabularnewline
\hline 
\toprule 
\multirow{3}{*}{{\rotatebox[origin=c]{90}{Random}}}
& $\mathsf{HalfCheetah}$  &---
&$13.1\pm 1.1$ 
& \colorbox{tacell!30}{$30.2$}$\pm1.3$
& \colorbox{tacell!30}{$30.1$}$\pm1.3$ \tabularnewline
 & $\mathsf{Hopper}$  &---
&$7.9\pm 0.2$  
&$7.5\pm0.9$
&$8.1\pm1.7$\tabularnewline
  &$\mathsf{Walker2d}$ &---

& \colorbox{tacell!30}{$5.4$}$\pm 1.2$
& \colorbox{tacell!30}{$5.9$}$\pm8.1$
&$4.5\pm4.9$ \tabularnewline
\hline
\multirow{3}{*}{\rotatebox[origin=c]{90}{\parbox[c]{1.2cm}{\centering Medium Replay}}}  & $\mathsf{HalfCheetah}$  &$36.6\pm0.8$
&$44.2\pm1.2$ 
&$42.2\pm0.7$
& \colorbox{tacell!30}{$54.4$} $\pm1.8$ \tabularnewline
  & $\mathsf{Hopper}$ &$82.7\pm7.0$

&\colorbox{tacell!30}{$94.7$}$\pm 8.6$
& \colorbox{tacell!30}{$92.6$}$\pm10.2$
&$68.1\pm19.7$  \tabularnewline
 & $\mathsf{Walker2d}$ &$66.6\pm3.0$
&$73.8\pm 7.1$
& \colorbox{tacell!30}{$83.7$}$\pm9.0$
&$80.0\pm11.4$  \tabularnewline
\hline
\multirow{3}{*}{{\rotatebox[origin=c]{90}{Medium}}} & $\mathsf{HalfCheetah}$ &$42.6\pm0.1$
&$47.4\pm0.2$
&$44.1\pm0.5$
& \colorbox{tacell!30}{$56.4$}$\pm0.8$  \tabularnewline
 & $\mathsf{Hopper}$ &$67.6\pm1.0$
&$66.2\pm 5.7$ 
& \colorbox{tacell!30}{$78.6$}$\pm15.7$
&$66.6\pm24.3$  \tabularnewline
  & $\mathsf{Walker2d}$ &$74.0\pm1.4$

&$78.3 \pm 8.7$
&$79.0\pm3.1$
& \colorbox{tacell!30}{$83.3$}$\pm3.5$
\tabularnewline
\hline
\multirow{3}{*}{\rotatebox[origin=c]{90}{\parbox[c]{1.2cm}{\centering Medium Expert}}}  & $\mathsf{HalfCheetah}$ &$86.8\pm1.3$
&$86.7\pm5.3$ 
& \colorbox{tacell!30}{$91.3$}$\pm0.9$
&$90.9\pm1.0$  \tabularnewline
 & $\mathsf{Hopper}$ &$107.6\pm1.8$

& \colorbox{tacell!30}{$91.5$}$\pm14.3$ 
&$56.9\pm21.3$
&$73.6\pm25.5$  \tabularnewline
  & $\mathsf{Walker2d}$  &$108.1\pm 0.2$

&\colorbox{tacell!30}{$110.1$} $\pm0.5$
&$105.2\pm6.1$
& \colorbox{tacell!30}{$107.5$}$\pm1.2$ \tabularnewline
\hline
\multirow{3}{*}{{\rotatebox[origin=c]{90}{Expert}}} & $\mathsf{HalfCheetah}$ &---

&\colorbox{tacell!30}{$95.0$}$\pm0.5$ 
&$93.8\pm1.0$
&$93.7\pm1.0$  \tabularnewline
 & $\mathsf{Hopper}$ &---

&\colorbox{tacell!30}{$109.4$}$\pm0.5$ 
&$95.0\pm21.1$
&$97.8\pm17.0$  \tabularnewline
  & $\mathsf{Walker2d}$ &---

&\colorbox{tacell!30}{$109.0$}$\pm 1.2$
&$107.0\pm0.6$
&$107.0\pm0.8$  \tabularnewline
\hline
\toprule
\multicolumn{2}{c}{Total (without random \& expert)} &$672.6$
& $692.8$ & $673.6$ & $680.8$ \tabularnewline
\hline
\multicolumn{2}{c}{Total } & ---
&\colorbox{tacell!30}{$1032.7$}
& \colorbox{tacell!30}{$1013.2$}
& \colorbox{tacell!30}{$1021.8$}
\tabularnewline
\hline
\toprule

\end{tabular}
}
\end{center}

	\label{tab:exp-result2} 
	\vspace{-5mm}

\end{table}

\subsection{Performance metric}\label{sec:metric}
With an output score of some method evaluated on MuJoCo in D4RL, we use the widely used {\em normalized performance score} as the performance metric:
  $\text{normalized score} \defn \frac{\text{score} - \text{random score}}{\text{expert score} - \text{random score}} \cdot 100,$
where the expert score (resp.~random score) represents the performance of an expert policy (resp.~a random policy). Here, noting that different offline datasets (such as {\em expert} dataset, {\em medium} dataset) for the same task enjoy a same expert/random score; for self-consistency, we recall the expert scores and random scores in different tasks in Table~\ref{tab:scores}.
\begin{table*}[ht]
\begin{center}
\begin{tabular}{c|cc}
\hline 
\toprule
  & {\bf expert score} & {\bf random score} \tabularnewline
\hline 
\toprule 
 $\mathsf{HalfCheetah}$ & $12135.0$ & $-280.18$
 \tabularnewline
 $\mathsf{Hopper}$ & $3234.3$ & $-20.27$
 \tabularnewline
$\mathsf{Walker2d}$ & $4592.3$ & $1.63$
 \tabularnewline
\hline
\toprule

\end{tabular}
\end{center}

	\caption{The expert scores and random scores of different MuJoCo tasks in D4RL \cite{fu2020d4rl}.}
	\label{tab:scores} 

\end{table*}


\subsection{Auxiliary implementation details}
For our two methods \our and \ourr, the models of the policies (actors) and the Q-functions (critics) are the same as the ones in {\sf TD3-BC} \cite{fujimoto2021minimalist}. The models for $\qsarsa$ (including the networks parameterized by $\theta_s$ and $\overline{\theta}_s$) are MLPs with ReLU activations and with 2 hidden layers of width 1024.
The training of the $\qsarsa$ network parameterized by $\theta_s$ is completed in the first training phase using Adam with initial learning rate $10^{-4}$ and batch size as $512$. The target of $\qsarsa$ is updated smoothly with $\tau = 0.005$, i.e., $\overline{\theta}_s \leftarrow (1-\tau)\overline{\theta}_s + \tau \theta_s$.

\end{document}